\documentclass[final]{cvpr}

\usepackage{times}
\usepackage{epsfig}
\usepackage{graphicx}
\usepackage{amsmath}
\usepackage{amssymb}

\usepackage{booktabs}
\usepackage{threeparttablex}
\usepackage{xspace}
\usepackage{multirow}
\usepackage{bbding}
\usepackage{multicol}
\usepackage{appendix}
\graphicspath{{figures/}}
\usepackage{color}
\usepackage{float}
\usepackage[caption=false]{subfig}
\usepackage{listings}
\usepackage{colortbl}
\usepackage{enumitem}
\usepackage{booktabs}
\usepackage{tablefootnote}
\definecolor{gray}{rgb}{0.6,0.6,0.6}

\usepackage[pagebackref=true,breaklinks=true,colorlinks,bookmarks=false]{hyperref} 
\newcolumntype{x}[1]{>{\centering\arraybackslash}p{#1pt}}

\newlength\savewidth\newcommand\shline{\noalign{\global\savewidth\arrayrulewidth
		\global\arrayrulewidth 1pt}\hline\noalign{\global\arrayrulewidth\savewidth}}
\newcommand{\tablestyle}[2]{\setlength{\tabcolsep}{#1}\renewcommand{\arraystretch}{#2}\centering\footnotesize}

\def\cvprPaperID{xxxx} 
\def\confYear{CVPR 2021}

\begin{document}

\title{TDN: Temporal Difference Networks for Efficient Action Recognition}

\author{Limin Wang \quad Zhan Tong \quad Bin Ji \quad Gangshan Wu \\
State Key Laboratory for Novel Software Technology, Nanjing University, China\\
{\tt\small 07wanglimin@gmail.com, tongzhan@smail.nju.edu.cn, binjinju@smail.nju.edu.cn, gswu@nju.edu.cn}  
}

\maketitle

\pagestyle{empty}
\thispagestyle{empty} 

\begin{abstract}
Temporal modeling still remains challenging for action recognition in videos. To mitigate this issue, this paper presents a new video architecture, termed as Temporal Difference Network (TDN), with a focus on capturing multi-scale temporal information for efficient action recognition. The core of our TDN is to devise an efficient temporal module (TDM) by explicitly leveraging a temporal difference operator, and systematically assess its effect on short-term and long-term motion modeling. To fully capture temporal information over the entire video, our TDN is established with a two-level difference modeling paradigm. Specifically, for local motion modeling, temporal difference over consecutive frames is used to supply 2D CNNs with finer motion pattern, while for global motion modeling, temporal difference across segments is incorporated to capture long-range structure for motion feature excitation. TDN provides a simple and principled temporal modeling framework and could be instantiated with the existing CNNs at a small extra computational cost. Our TDN presents a new state of the art on the Something-Something V1 \& V2 datasets  and is on par with the best performance on the Kinetics-400 dataset. In addition, we conduct in-depth ablation studies and plot the visualization results of our TDN, hopefully providing insightful analysis on temporal difference modeling. We release the code at \href{https://github.com/MCG-NJU/TDN}{https://github.com/MCG-NJU/TDN}.
\end{abstract}

\section{Introduction}

Deep neural networks have witnessed great progress for action recognition in videos~\cite{large_scale_cnn,DBLP:conf/nips/SimonyanZ14,DBLP:conf/eccv/WangXW0LTG16,DBLP:conf/iccv/TranBFTP15,DBLP:journals/corr/abs-1812-03982,LGD,Wang0T15}. Temporal modeling is crucial for capturing motion information in videos for action recognition, and this is usually achieved by two kinds of mechanisms in the current deep learning approaches. One common method is to use a two-stream network~\cite{DBLP:conf/nips/SimonyanZ14}, where one stream is on RGB frames to extract appearance information, and the other is to leverage optical flow as an input to capture movement information. This method turns out to be effective for improving action recognition accuracy but requires high computational consumption for optical flow calculation. Another alternative approach is to use 3D convolutions ~\cite{DBLP:conf/icml/JiXYY10,DBLP:conf/iccv/TranBFTP15} or temporal convolutions~\cite{DBLP:conf/cvpr/TranWTRLP18,DBLP:conf/eccv/XieSHTM18,Qiu2017LearningSR} to implicitly learn motion features from RGB frames. However, 3D convolutions often lack specific consideration in the temporal dimension and might bring higher computational cost as well. Therefore, designing an effective temporal module of high motion modeling power and low computational consumption is still a challenging problem for video recognition.

\begin{figure}
    \centering
    \includegraphics[width=0.9\linewidth]{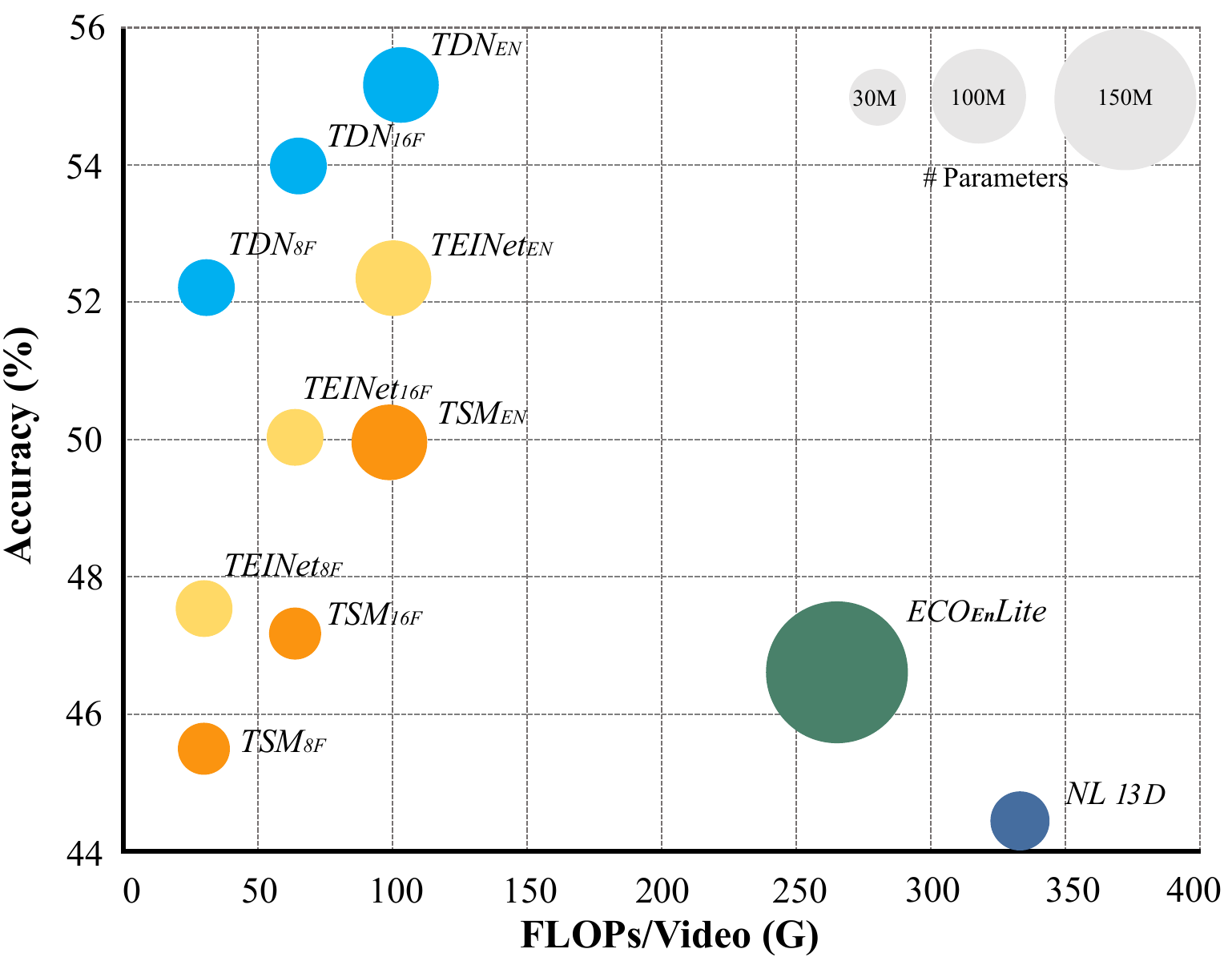}
    \caption{\small Video classification performance comparison on Something-Something V1~\cite{sth} in terms of Top1 accuracy, computational cost, and model size. Our proposed TDN achieves the best trade-off between accuracy and efficiency, when compared with previous methods such as NL I3D~\cite{i3d_gcn}, ECO~\cite{eco}, TSM~\cite{DBLP:conf/iccv/LinGH19} and TEINet~\cite{DBLP:journals/corr/abs-1911-09435}.}
    \label{fig:motivation}
    \vspace{-5mm}
\end{figure}

This paper aims to present a new temporal modeling mechanism by introducing a temporal difference based module (TDM). Temporal derivative (difference) is highly relevant with optical flow~\cite{DBLP:journals/ai/HornS81}, and has shown effectiveness in action recognition by using RGB difference as an approximate motion representation~\cite{DBLP:conf/eccv/WangXW0LTG16,ZhaoXL18}. However, these approaches simply treat RGB difference as another video modality and train a different network to fuse with the RGB network. Instead, we aim to present a unified framework to capture appearance and motion information jointly, by generalizing the idea of temporal difference into a principled and efficient temporal module for end-to-end network design.  

In addition, we argue that both short-term and long-term temporal information are crucial for action recognition, in the sense that they are able to capture the distinctive and complementary properties of an action instance. Therefore, in our proposed temporal modeling mechanism, we present a unique two-level temporal modeling framework based on a holistic and sparse sampling strategy~\cite{DBLP:conf/eccv/WangXW0LTG16}, termed as Temporal Difference Network (TDN). Specifically, in TDN, we consider two efficient forms of TDMs for motion modeling at different scales. For local motion modeling, we present a light-weight and low-resolution difference module to supply a single RGB with motion patterns via lateral connections, while for long-range motion modeling, we propose a multi-scale and bidirectional difference module to capture cross-segment variations for motion excitation. These two TDMs are systematically studied as modular building blocks for short-term and long-rang temporal structure extraction.

Our TDN provides a simple and general video-level motion modeling framework and could be instantiated with existing CNNs at a small extra computational cost. To demonstrate the effectiveness of TDN, we implement it with ResNets and perform experiments on two datasets: Kinetics and Something-Something. The evaluation results show that our TDN is able to yield a new state-of-the-art performance on both motion relevant Something-Something dataset and scene relevant Kinetics dataset, under the setting of using similar backbones. As shown in Figure~\ref{fig:motivation}, our best result is significantly better than previous methods on the dataset of Something-Something V1. We also perform detailed ablation studies to demonstrate the importance of temporal difference operation and investigate the effect of a specific design of TDM. In summary, our main contribution lies in the following three aspects:
\vspace{-1mm}
\begin{itemize}
\item We generalize the idea of RGB difference to devise an efficient temporal difference module (TDM) for motion modeling in videos and provide an alternative to 3D convolutions by systematically presenting effective and detailed module design.
\vspace{-2mm}
\item Our TDN presents a video-level motion modeling framework with the proposed temporal difference module, with a focus on capturing both short-term and long-term temporal structure for video recognition.
\vspace{-2mm}
\item Our TDN obtains the new state-of-the-art performance on the datasets of Kinetics and Something-Something under the setting of using the same backbones. We also perform in-depth ablation study on TDM to provide some insights on our temporal difference modeling.
\end{itemize}

\vspace{-1.5mm}
\section{Related work}

\paragraph{\bf Short-term temporal modeling.} Action recognition has attracted lots of research attention in the past few years. These methods could be categorized into two types: (1) two-stream CNNs~\cite{DBLP:conf/nips/SimonyanZ14} or its variants~\cite{DBLP:conf/cvpr/FeichtenhoferPZ16}: it used two inputs of RGB and optical flow to separately model appearance and motion information in videos with a late fusion; (2) 3D-CNNs~\cite{DBLP:conf/iccv/TranBFTP15,DBLP:conf/icml/JiXYY10}: it proposed 3D convolution and pooling to directly learn spatiotemporal features from videos. Several variants tried to reduce the computation cost of 3D convolution by decomposing it into a 2D convolution and a 1D temporal convolution, for example R(2+1)D~\cite{DBLP:conf/cvpr/TranWTRLP18}, S3D~\cite{DBLP:conf/eccv/XieSHTM18}, P3D~\cite{Qiu2017LearningSR}, and CT-Net~\cite{li2021ctnet}. Following this research line, several works focused on designing more powerful temporal modules and inserted them into a 2D CNN for efficient action recognition, such as TSM~\cite{DBLP:conf/iccv/LinGH19}, TIN~\cite{tin}, TEINet~\cite{DBLP:journals/corr/abs-1911-09435}, TANet~\cite{TANet}, and TEA~\cite{li2020tea}. In addition, some methods tried to leverage the idea of two stream network to design a multi-branch architecture to capture both appearance and motion or context information, with a carefully designed temporal module or two RGB inputs sampled at different FPS, including Non-local Net~\cite{nonlocal}, ARTNet~\cite{DBLP:conf/cvpr/WangL0G18}, STM~\cite{DBLP:conf/iccv/JiangWGWY19}, SlowFast~\cite{DBLP:journals/corr/abs-1812-03982}, and CorrelationNet~\cite{CorrNet}. Some recent works~\cite{x3d} tried network architecture search for video recognition. These works were clip-based architecture with a focus on short-term motion modeling by learning from a small portion of the whole video (e.g., 64 frames).

\vspace{-4mm}
\paragraph{\bf Long-term temporal modeling.} Short-term clip-based networks fail to capture the long-range temporal structure. Several methods were proposed to overcome this limitation by stacking more frames with RNN~\cite{Ng15,DonahueJ2015} or long temporal convolution~\cite{DBLP:journals/pami/VarolLS18}, or using a sparse sampling and aggregation strategy~\cite{DBLP:conf/eccv/WangXW0LTG16,DBLP:conf/eccv/ZhouAOT18,zhang2019v4d,DBLP:conf/aaai/HeZGLLLWW19,li2020tea}. Among these methods, temporal segment network (TSN)~\cite{DBLP:conf/eccv/WangXW0LTG16} turned out to be an effective long-range modeling framework and obtained the state-of-the-art performance with 2D CNNs on several benchmarks. However, TSN with 2D CNNs only performed temporal fusion at the last stage and failed to capture the finer temporal structure. StNet~\cite{DBLP:conf/aaai/HeZGLLLWW19} proposed a local and global module to model temporal information hierarchically.  V4D~\cite{zhang2019v4d} extended the TSN framework by proposing a principled 4D convolutional operator to aggregate long-range information from different stages.

\begin{figure*}[t]
    \begin{center}
    \includegraphics[width=0.9\textwidth]{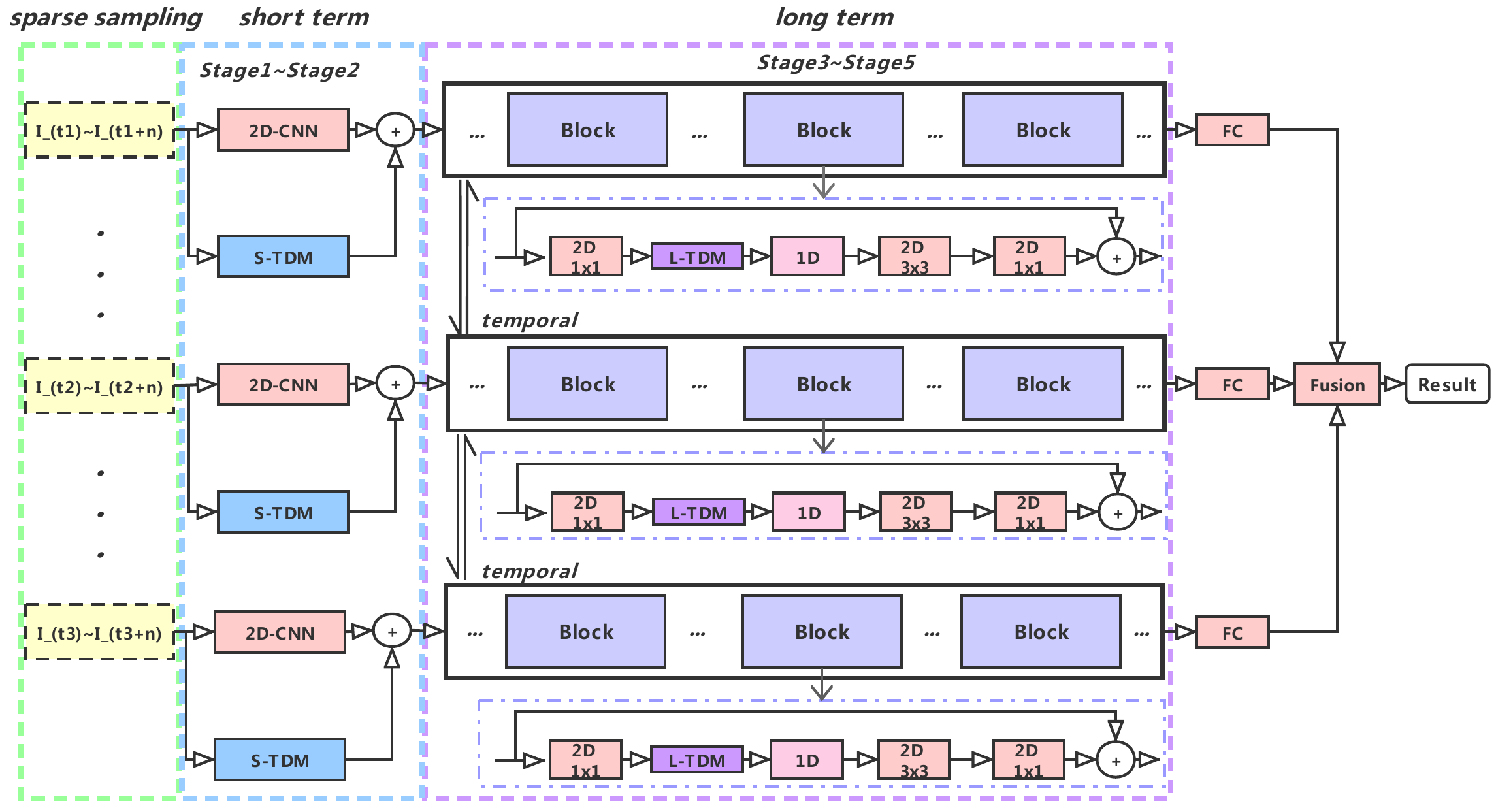}
    \end{center}
    \vspace{-3mm}
    \caption{\small {\bf Temporal Difference Network.} We present a video-level framework for learning action models from the entire video, coined as TDN. Based on the sparse sampling from multiple segments, our TDN aims to model both short-term and long-term motion information in our framework. The key contribution is to design an efficient short-term temporal difference module (S-TDM) and a long-term temporal difference module (L-TDM), to supply a 2D CNN with local motion information and enable long-range modeling across segments, respectively. CNNs share the same parameters on all segments.  Details on both modules could be found in Figure~\ref{fig:module}.}
    \label{fig:framework}
    \vspace{-2mm}
\end{figure*}

\vspace{-4mm}
\paragraph{\bf Temporal difference representation.} Temporal difference operations appeared in several previous works for motion extraction, such as RGB Difference~\cite{DBLP:conf/eccv/WangXW0LTG16,ZhaoXL18,tdiff} and Feature Difference~\cite{DBLP:journals/corr/abs-1911-09435, DBLP:conf/iccv/JiangWGWY19,li2020tea}. RGB difference turned out to be an efficient alternative modality to optical flow as motion representation~\cite{DBLP:conf/eccv/WangXW0LTG16,ZhaoXL18,tdiff}. However, they only treated RGB differently with another video modality and trained a separate network to fuse with RGB stream. The work of TEINet~\cite{DBLP:journals/corr/abs-1911-09435}, TEA~\cite{li2020tea}, and STM~\cite{DBLP:conf/iccv/JiangWGWY19} employed a difference operation for network design. However, these two methods simply used a simple difference operator for single-level motion extraction and received less research attention than 3D convolutions.

Different from the existing methods, our proposed temporal difference network (TDN) is a video-level architecture of capturing both short-term and long-term information for end-to-end action recognition. Our key contribution is to introduce a temporal difference module (TDM) to explicitly compute motion information, and efficiently leverage it into our two-level motion modeling paradigm. We hope to improve and popularize these temporal difference-based modeling alternatives, which turns out to generally outperform 3D convolutions on two benchmarks with smaller FLOPs.

\section{Temporal Difference Networks}
In this section, we describe our Temporal Difference Network (TDN) in detail. First, we give an overview of the TDN framework, which is composed of a short-term and long-term temporal difference module (TDM). Then, we give a technical description of both modules. Finally, we provide the implementation detail to instantiate TDN with a ResNet backbone.

\subsection{Overview}

As shown in Figure~\ref{fig:framework}, our proposed temporal difference network (TDN) is a video-level framework for learning action models by using the entire video information. Due to the limit of GPU memory, following TSN framework~\cite{DBLP:conf/eccv/WangXW0LTG16}, we present a sparse and holistic sampling strategy for each video. Our key contribution is to leverage the temporal difference operator into the network design to explicitly capture both short-term and long-term motion information. Efficiency is our core consideration in temporal difference module (TDM) design, and we investigate two specific forms to accomplish the tasks of motion supplement in a local window and motion enhancement across different segments, respectively. These two modules are incorporated into the main network via a residual connection.

\begin{figure*}[t] 
    \begin{center}
    \includegraphics[width=1\textwidth]{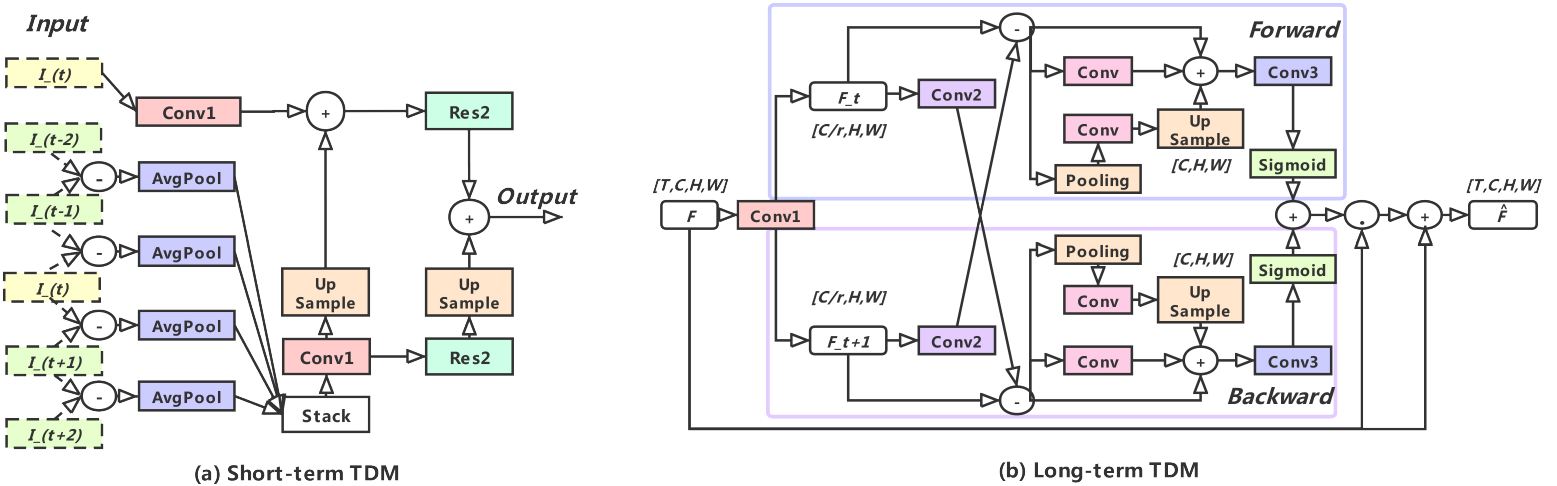}
    \end{center}
    \vspace{-2mm}
    \caption{An illustration of the short-term TDM and long-term TDM. {\em Left}: our S-TDM operates on the stacked RGB difference and is fused with single RGB CNN via residual connection to capture the short-term motion. {\em Right}: our L-TDM presents a bi-directional and multi-scale attention mechanism to leverage cross-segment information to enhance the frame-wise representations. More details could be found in the text.}
    \label{fig:module}
    \vspace{-2mm}
\end{figure*}

Specifically, each video $V$ is divided into $T$ segments of equal duration without overlapping. We randomly sample a frame from each segment and totally obtain $T$ frames $\mathbf{I} = [I_1, \cdots, I_T]$, where the shape of $\mathbf{I}$ is $[T, C, H, W]$. These frames are separate fed into a 2D CNN to extract frame-wise features $\mathbf{F} = [F_1, \cdots, F_T]$, where $\mathbf{F}$ denotes the feature representation in the hidden layer and its dimension is $[T, C', H', W']$. The short-term TDM aims to supply these frame-wise representation $\mathbf{F}$ of early layers with local motion information to improve its representation power:
\begin{equation}
    \mathrm{\mathbf{Short \ term \ TDM}:} \ \ \ \ \hat{F}_i =  F_i + \mathcal{H}(I_i),
\end{equation}
where $\hat{F}_i$ denotes the enhanced representation by TDM, $\mathcal{H}$ denotes our short-term TDM, and it extracts local motion from adjacent frames around $I_i$. The long-term TDM aims at leveraging cross-segment temporal structure to enhance frame-level feature representation:
\begin{equation}
    \mathrm{\mathbf{Long \ term \ TDM}:} \ \ \ \ \hat{F}_i =  F_i + F_i \odot \mathcal{G}(F_{i}, F_{i+1}),
    \label{equ:ltdm}
\end{equation}
where $\mathcal{G}$ represents our long-term TDM, and in the current implementation, we only consider adjacent segment-level information for long-range temporal modeling in each long-term TDM. By stacking multiple long-term TDMs, we are able to capture temporal structure over a long scale. Details will be described in the next subsections.

\subsection{Short-term TDM}

We argue that adjacent frames are very similar in a local temporal window and directly stacking multiple frames for subsequent processing is inefficient. On the other hand, sampling a single frame from each window is able to extract appearance information, but fails to capture local motion information. Therefore, our short-term TDM chooses to supply a single RGB frame with a temporal difference to yield an efficient video representation, explicitly encoding both appearance and motion information.

Specifically, our short-term TDM operates at early layers of networks for low-level feature extraction and enables a single frame RGB to be aware of local motion via fusing temporal difference information. As shown in Figure~\ref{fig:module}, for each sampled frame $I_i$, we extract several temporal RGB difference in a local window centered at $I_i$, and then stack them along channel dimension $\mathbf{D}(I_i) = [D_{-2}, D_{-1}, D_1, D_2]$. Based on this representation, we present an efficient form of TDM:
\begin{equation}
    \mathcal{H}(I_i) =  \mathrm{Upsample} (\mathrm{CNN}( \mathrm{Downsample} (\mathbf{D}(I_i)) ) ),
\end{equation}
where $D$ represents the RGB difference around $I_i$, and $\mathrm{CNN}$ is the specific network for different stages. To keep the efficiency, we design a light-weight CNN module to operate on the stacked RGB difference $\mathbf{D}(I_i)$. It generally follows a low-resolution processing strategy: (1) downsample RGB difference by half with an average pooling, (2) extract motion features with a 2D CNN, (3) upsample motion features to match RGB features. This design comes from our observation that RGB difference exhibits very small values for most areas and only contains high response in motion salient regions. So, it is enough to use low-resolution architecture for this sparse signal without much loss of accuracy.

The information of short-term TDM is fused with the single RGB frame, so that the original frame-level representation is aware of motion pattern and able to better describe a local temporal window. We implement this fusion with lateral connections. We attach a fusion connection from short-term TDM to frame-level representation for each early stage (i.e., Stage 1-2 in our experiments). In practice, we also compare the residual connection with other fusion strategies as shown in the ablation study.

\subsection{Long-term TDM}
The frame-wise representation equipped with short-term TDM is powerful for capturing spatiotemporal information within a local segment (window). However, this representation is limited in terms of the temporal receptive field and thus fails to explore the long-range temporal structure for learning action models. Thus, our long-term TDM tries to use cross-segment information to enhance the original representation via a novel bidirectional and multi-scale temporal difference module.

In additional to efficiency, the missing-alignment of spatial location between long-range frames is another issue. Consequently, we devise a multi-scale architecture to smooth difference in large a receptive field before difference calculation.  As shown in Figure~\ref{fig:module}, we first compress the feature dimension by a ratio $r$  with a convolution for efficiency, and calculate the aligned temporal difference through adjacent segments:
\begin{equation}
    C(F_i, F_{i+1}) = F_i -  \mathrm{Conv} (F_{i+1}),
\end{equation}
where $C(F_i, F_{i+1})$ represents the aligned temporal difference for segment $F_i$, $\mathrm{Conv}$ is the channel-wise convolution for spatially smoothing and thus relieving the missing-alignment issue. Then, the aligned temporal difference undergoes through a multi-scale module for long-range motion information extraction:
\begin{equation}
\small
M(F_i, F_{i+1}) = \mathrm{Sigmd}(\mathrm{Conv}(\sum_{j=1}^N \mathrm{CNN}_j (C(F_i, F_{i+1})))),
\end{equation}
where $\mathrm{CNN}_j$ at different spatial scales aims at extracting motion information from different receptive field, and $N=3$ in practice. Their fusion could be more robust for the missing-alignment issue. In implementation, it involves three branches: (1) short connection, (2) a 3$\times$3 convolution, and (3) a average pooling, a 3$\times$3 convolution, and a bilinear upsampling. Finally, we utilize bidirectional cross-segment temporal difference to enhance fame level features as follows:
\begin{equation}
\small
    F_i \odot \mathcal{G}(F_{i}, F_{i+1}) = F_i \odot  \frac{1}{2}[M(F_i, F_{i+1}) + M(F_{i+1}, F_{i})],
    \label{equ:ltdm1}
\end{equation}
where $\odot$ is the element-wise multiplication. We also combine the original frame level representation and enhance representation via a residual connection as in Eq. (\ref{equ:ltdm}). Slightly different from short-term TDM, we employ the difference representation as an attention map to enhance frame level features, which is partially based on the observation that attention modeling is more effective for latter stage of CNNs. We also compare this implementation with other forms in the ablation study.

\begin{table*}[t]\centering\vspace{-1.5em}
		\captionsetup[subfloat]{captionskip=2pt}
		\captionsetup[subffloat]{justification=centering}
		\subfloat[\textbf{Study on the effect of difference operator}: We compare with baselines by directly stacking or averaging temporal frames, which are worse than temporal difference operator for both short-term and long-term modeling.
		\label{tbl:ablation_diff}]{
			\tablestyle{2pt}{1.05}
                \begin{tabular}{c|c|x{30}x{30}}
                ~~S-TDM~~ & ~~L-TDM~~ &  FLOPs  & Top1 \\
                \shline
                concat & avg & 36.2G & 41.5\%  \\
                concat & diff\checkmark & 36.2G & 51.4\%  \\
                diff\checkmark & avg & 35.9G & 51.6\%  \\
                \hline
                diff\checkmark & diff\checkmark & {\bf 35.9G} & {\bf 52.3\%} \\
                \multicolumn{4}{c}{} \\ 
                \end{tabular}}
                \hspace{5mm}
		\subfloat[\textbf{Study on S-TDM}: We compare different implementation forms of S-TDM, including spatiotemporal attention, channel attention, combination of residual connection and attention.
		\label{tbl:stdm}]{
			\tablestyle{2pt}{1.05}
			 \begin{tabular}{c|x{30}}
                  Fusion &  Top1 \\
                 \shline
                   $F \odot \mathcal{H}$ & 43.7\%\\
                   $F + F \odot \mathcal{H}$  & 47.6\% \\
                   $F + \mathcal{H}$ & {\bf 52.3\%} \\
                   \hline
                  $ F \odot \mathcal{H}_{channel}$  & 47.3\% \\
                  ~~~~~$F + F \odot \mathcal{H}_{channel}$ ~~~~~~ &  47.9\% \\
                  \end{tabular}}\hspace{5mm}
		\subfloat[\textbf{Study on L-TDM}: We compare different implementation forms of L-TDM, with a focus on multi-scale representation and bidirectional difference, and different fusion strategies, including residual connection, channel attention, and spatiotemporal attention.
		\label{tbl:ltdm} ]{
			\tablestyle{2pt}{1.05}
			\begin{tabular}{c|cc|x{30}}
                  Fusion & Multi-scale &  bidirect. &Top1 \\
                 \shline
                  $F + \mathcal{G}$ & \checkmark & \checkmark & 44.1\%  \\
                  $F + F \odot \mathcal{G}_{channel}$&  & \checkmark & 50.9\% \\
                  \hline
                  $F + F \odot \mathcal{G}$ & \checkmark & & 50.0\%  \\
                  $F + F \odot \mathcal{G}$ &  & \checkmark& 49.7\%  \\
                  \hline
                  $F + F \odot \mathcal{G}$ & \checkmark & \checkmark & {\bf 52.3\%} \\
                  \end{tabular}}
\vspace{-6 mm}
        \subfloat[\textbf{Study on the location of S-TDM and L-TDM}: We place S-TDMs and L-TDMs at different stages of ResNet50. The results imply that it obtains the best performance when stages 1-2 focus on short-term modeling and stages 3-5 focus on long-term modeling.
        \label{tbl:ablation_loc}]{
		\tablestyle{2pt}{1.05}
        \begin{tabular}{c|c|x{30}x{30}}
        \multicolumn{4}{c}{} \\ 
        S-TDM &   L-TDM & FLOPs & Top1 \\
        \shline
         - & - & 33G & 46.6\% \\
         Stage 1 & Stage 2-5 & 35G & 50.6\% \\
         Stage 1-2 & Stage 3-5 & 36G & {\bf 52.3\%} \\
         Stage 1-3 & Stage 4-5 & 38G & 51.7\% \\
         	\multicolumn{4}{c}{} \\ 
				\multicolumn{4}{c}{} \\ 
				\multicolumn{4}{c}{} \\ 
        \end{tabular}} \hspace{6mm}
        \subfloat[\textbf{S-TDM vs. L-TDM}: We compare the effectiveness of S-TDM and L-TDM. The results imply that S-TDM is slightly better than L-TDM when simply using a single types of TDM, and S-TDM and L-TDM are complementary to each other. 
        \label{tbl:comp1}]{
		\tablestyle{2pt}{1.05}
        \begin{tabular}{cc|ccc|x{30}}
        \multicolumn{2}{c|}{S-TDM}& \multicolumn{3}{c|}{L-TDM} & \multirow{2}{*}{Top1} \\
          Conv1 & Res2 & Res3 & Res4 & Res5 & \\
          \shline
           &  & & & & 46.6\% \\
          \checkmark &  & & & & 51.3\% \\
          \checkmark & \checkmark & & & & 51.5\%  \\
          \hline
           & &\checkmark & \checkmark & \checkmark & 49.9\%    \\
          \checkmark & \checkmark & \checkmark &\checkmark & \checkmark & {\bf 52.3\% } \\
		  \multicolumn{6}{c}{} \\ 
		  \multicolumn{6}{c}{} \\ 
          \end{tabular}}\hspace{8mm}
        \subfloat[\textbf{Comparison with other temporal modules}:  We compare TDM with several temporal modules: Temporal convolution, TSM, and TEINet. For fair comparison, we report the result with 8 frame and 40 for each temporal module (++ for 40 frames). Our TDM is better than previous temporal modules.
        \label{tbl:modulecmp}]{
		\tablestyle{2pt}{1.05}
        \begin{tabular}{c|x{35}x{30}x{30}}
        \multicolumn{4}{c}{} \\ 
        {Model}  & FLOPs & {Top1} & {Top5}\\
        \shline
        T-Conv~\cite{DBLP:conf/cvpr/TranWTRLP18} & 33G & 47.5\% & 77.5\% \\
        T-Conv++~\cite{DBLP:conf/cvpr/TranWTRLP18} & 165G & 48.2\% & 79.1\% \\
        TSM~\cite{DBLP:conf/iccv/LinGH19} &  33G & 47.1\% & 76.2\% \\
        TSM++~\cite{DBLP:conf/iccv/LinGH19} &  165G & 47.6\% & 77.9\% \\
        TEINet~\cite{DBLP:journals/corr/abs-1911-09435} & 33G & 48.4\% & 77.2\% \\
        TEINet++~\cite{DBLP:journals/corr/abs-1911-09435} & 165G & 49.0\% & 79.0\% \\
        \hline
        TDM & 36G & {\bf 52.3\%} & {\bf 80.6\%} \\
        \end{tabular}}
		\caption{\small {Ablations} on \textbf{Something-Something V1} with 8-frame TDN-ResNet50. We show top-1 classification accuracy (\%), and computational cost measured in FLOPs (floating-point operations, in \# of multiply-adds ) for a 1-clip and center-crop input of size 224$\times$224.} 
 \vspace{-2mm}
		\label{tab:ablations}
\end{table*}

\subsection{Exemplar: TDN-ResNet}

As discussed above, our TDN framework is based on sparse sampling of TSN~\cite{DBLP:conf/eccv/WangXW0LTG16}, which operates on a sequence of frames uniformly distributed over the entire video. Our TDN presents a two-level motion modeling mechanism, with a focus on capturing temporal information in a local-to-global fashion. In particular, as shown in Figure~\ref{fig:framework}, we insert short-term TDMs (S-TDM) in early stages for finer and low-level motion extraction, and long-term TDMs (L-TDM) into latter stages for coarser and high-level temporal structure modeling.

We instantiate our TDN with a ResNet backbone~\cite{resnet}. Following the practice in V4D~\cite{zhang2019v4d}, the first two stages of ResNet are for short-term temporal information extraction within each segment by using S-TDMs, and the latter three stages of ResNet are equipped with L-TDMs for capturing long-range temporal structure across segments. For local motion modeling, we add both residual connections between S-TDM and the main network for Stage 1 and Stage 2. For long term motion modeling, we add L-TDM and a temporal convolution in each residual block of Stages 3-5. In practice, the final TDN-ResNet only increases the FLOPs over the original 2D TSN-ResNet by around 9\%.

\section{Experiments}

In this section, we present the experiment results of our TDN framework. First, we describe the evaluation datasets and implementation details. Then, we perform ablation studies on the design of our TDN. Next, we compare our TDN with the state-of-the-art methods. Finally, we show some visualization results to further analyze our TDN.

\subsection{Datasets and implementation details}

\noindent {\bf Video datasets.} We evaluate our TDN on two video datasets, which pay attention to different aspects of an action instance for recognition. {\bf Kinetics-400}~\cite{kinetics400} is a large-scale YouTube video dataset and has around 300k trimmed videos covering 400 categories. The Kinetics dataset contains activities in daily life and some categories are highly correlated with interacting objects or scene context. We train our TDN on the training data (around 240k videos) and report performance on the validation data (around 20k videos).  {\bf Something-Something}~\cite{sth} is a large-scale dataset created by crowdsourcing. The videos are collected by performing the same action with different objects so that action recognition is expected to focus on the motion property instead of objects or scene context. The first version contains around 100k videos over 174 categories, while the second version is with more videos, containing around 169k videos in training set and 25k videos in validation set. We report performance on the validation set of Something-Something V1 \& V2.

\noindent {\bf Training and testing.} In experiments, we use ResNet50 and ResNet101 to implement our TDN framework, and sample $T=8$ or $T=16$ frames from each video. Following common practice~\cite{DBLP:journals/corr/abs-1812-03982,nonlocal}, during training, each video frame is resized to have shorter side in $[256, 320]$ and a crop of $224 \times 224$ is randomly cropped. We pre-train our TDN on the ImageNet dataset~\cite{imagenet}. The batch size is 128 and the initial learning rate is 0.02. The total training epoch is set as 100 in the Kinetics dataset and 60 in the Something-Something dataset. The learning rate will be divided by a factor of 10 when the performance on validation set saturates. For testing, the shorter side of each video is resized to 256. We implement two kinds of testing scheme: {\bf 1-clip and center-crop} where only a center crop of $224 \times 224$ from a single clip is used for evaluation, and {\bf 10-clip and 3-crop} where three crops of $256 \times 256$ and 10 clips are used for testing. The first testing scheme is with high efficiency while the second one is for improving accuracy with a denser prediction scheme.

\subsection{Ablation studies}

We perform ablation studies on the Something-Something V1 dataset. For these evaluations, we use the testing scheme of 1 clip and center crop, and report the Top1 accuracy. We also compare with other temporal modeling modules to demonstrate the effectiveness of TDM.

\vspace{-4mm}
\paragraph{\bf Study on the effect of difference operation.} We begin our oblation study by exploring the effectiveness of temporal difference operation in our TDM. We implement fairly comparable baselines by simply removing temporal difference operation in S-TDM and replacing temporal difference with taking average in L-TDM. Table~\ref{tbl:ablation_diff} shows the results of various settings with temporal difference or without temporal difference. It can be seen that simply stacking and taking average to fuse temporal information will greatly decrease the recognition accuracy by around 10\%. We analyze that these temporal fusion strategies without difference operation would make the network to over-fit static information and fail to capture temporal variation in videos. Adding temporal difference in both S-TDM and L-TDM contributes to better accuracy and their combination obtains the best performance.

\vspace{-4mm}
\paragraph{\bf Study on short-term TDM.} We compare different forms of short-term TDM (S-TDM). We add long-term TDM (L-TDM) for all latter stages and place variations of S-TDM in early stages. As shown in Table~\ref{tbl:stdm}, we first compare different fusion strategies to combine difference representation with RGB features in S-TDM: (1) attention with element-wise multiplication, (2) addition with attention, (3) only addition. We can see that our S-TDM with simply addition yields the best performance and the other attention-based fusion might destroy the pre-trained feature correspondence. In addition, we try to use RGB difference representation to learn the channel attention weight just as TEINet~\cite{TANet}, and its performance is also worse than our proposed S-TDM (47.3\% vs. 52.3\%).  In the remaining study, we use the addition form of S-TDM by default.

\begin{table}[t]
\centering
\small
\tablestyle{1.6pt}{1}
\begin{tabular}{l|x{50}|x{24}|x{34}|x{20}|x{20}}
\multicolumn{1}{c|}{\multirow{2}*{\bf Method}} & \multirow{2}*{\bf Backbone} & \multirow{2}*{\bf Frames} & \multirow{2}*{\bf GFLOPs} & \multicolumn{2}{c}{\bf Sth-Sth V1}  \\
 &  &  &  &   {\bf Top1} & {\bf Top5} \\
\shline
TSN-RGB~\cite{DBLP:conf/eccv/WangXW0LTG16}& BNInception& 8 &16 & 19.5 & - \\
TRN-Multiscale~\cite{DBLP:conf/eccv/ZhouAOT18}& BNInception &  8 & 33 & 34.4 & -  \\
S3D-G~\cite{DBLP:conf/eccv/XieSHTM18} & Inception & $64$ & 71.38 & 48.2  & 78.7  \\
\hline \hline
TSM~\cite{DBLP:conf/iccv/LinGH19} & ResNet50 & $8$+$16$ & 98 & 49.7  & 78.5 \\
TEINet~\cite{DBLP:journals/corr/abs-1911-09435} & ResNet50 & $8$+$16$ & 99 & 52.5  & -  \\
TANet~\cite{TANet} & ResNet50 & $8$+$16$ & 99 & 50.6 & 79.3 \\
TEA~\cite{li2020tea} & ResNet50  &  16 & 70 & 51.9  &80.3  \\
TAM~\cite{tam} &bLResNet50 & 16$\times$2 & 47.7 & 48.4  & 78.8  \\
\hline \hline

ECO$_{En}$Lite~\cite{eco} & BNIncep+R18& $92$ & 267 & 46.4  & -  \\
I3D~\cite{I3D}& ResNet50 & 32$\times$2 & 306 & 41.6  & 72.2   \\
NL I3D+GCN~\cite{i3d_gcn}& R50+GCN &  32$\times$2 & 606 & 46.1  & 76.8  \\
GST~\cite{gst} & ResNet50 & 16 & 59 & 48.6  & 77.9    \\
STM~\cite{DBLP:conf/iccv/JiangWGWY19} & ResNet50 & 16$\times$30 & 67$\times$30 & 50.7  & 80.4   \\
V4D~\cite{zhang2019v4d} & ResNet50 & 8$\times$4 & 167.6 & 50.4   & -  \\
SmallBigNet~\cite{smallbignet}& ResNet50 & $8$+$16$ & 157 & 50.4  & 80.5   \\
CorrNet~\cite{CorrNet}& ResNet50 & 32$\times$10 & 115$\times$10 & 49.3  & -  \\
\hline \hline
 TDN (Ours) & ResNet50 &  8 & 36 & 52.3  & 80.6   \\
 TDN (Ours) & ResNet50 &  16 & 72 & 53.9  & 82.1    \\
 TDN (Ours) & ResNet50 &  8+16 & 108 & \textbf{55.1 } & \textbf{82.9 }  \\
 \hline \hline
CorrNet~\cite{CorrNet}  & ResNet101 & 32$\times$30 & 224$\times$30 & 51.7 & - \\
CorrNet~\cite{CorrNet}~\footnote{Pre-trained on Sports1M.}  & ResNet101 & 32$\times$30 & 224$\times$30 & 53.3 & - \\
GSM~\cite{GSM} & Inception V3 & fusion & 268 & 55.2 & -\\ 
\hline \hline
 TDN (Ours) & ResNet101 &  8 & 66 & 54.1  & 81.9   \\
 TDN (Ours) & ResNet101 &  16 & 132 & 55.3  &  83.3  \\
 TDN (Ours) & ResNet101 &  8+16 & 198 & \textbf{56.8 } & \textbf{84.1} \\
\multicolumn{1}{c|}{\multirow{2}*{\bf Method}} & \multirow{2}*{\bf Backbone} & \multirow{2}*{\bf Frames} & \multirow{2}*{\bf GFLOPs}  & \multicolumn{2}{c}{\bf Sth-Sth V2}  \\
 &  &  &  & {\bf Top1} & {\bf Top5} \\
\shline
TRN-Multiscale~\cite{DBLP:conf/eccv/ZhouAOT18}& BNInception &  8 & 33 & 48.8 & 77.6 \\
TAM~\cite{tam}  &bLResNet50 & 16$\times$2 & 47.7  & 61.7  & 88.1  \\
TSM~\cite{DBLP:conf/iccv/LinGH19} & ResNet50 & 16$\times$6 & 65$\times$6 & 63.4 & 88.5 \\
GST~\cite{gst} & ResNet50 & 16 & 59 & 62.6  & 87.9  \\
STM~\cite{DBLP:conf/iccv/JiangWGWY19} & ResNet50 & 16$\times$30 & 67$\times$30  & 64.2  & 89.8  \\
SmallBigNet~\cite{smallbignet}& ResNet50 & $8$+$16$ & 157   & 63.3  & 88.8  \\
TEINet~\cite{DBLP:conf/iccv/LinGH19} & ResNet50 & $8$+$16$ & 98 & 65.5 & 89.8 \\
TANet~\cite{TANet} & ResNet50 & 24$\times$6 & 99 $\times$6 & 66.0 & 90.1 \\
\hline \hline
 TDN (Ours) & ResNet50 &  8 & 36 & 64.0  & 88.8  \\
 TDN (Ours) & ResNet50 &  16 & 72 & 65.3  & 89.5  \\
 TDN (Ours) & ResNet50 &  8+16 & 108 &  {\bf 67.0} & {\bf 90.3 } \\
\hline \hline
 TDN (Ours) & ResNet101 &  8 & 66 & 65.8 & 90.2 \\
 TDN (Ours) & ResNet101 &  16 & 132 & 66.9 & 90.9  \\
 TDN (Ours) & ResNet101 &  8+16 & 198 & {\bf 68.2} & {\bf 91.6} \\
\end{tabular}
\vspace{0.5mm}
\caption{\small {\bf Comparison with the state-of-the-art methods on Something-Something V1 and V2}.  We instantiate our TDN with the backbones of ResNet50 and ResNet101 for evaluation. We compare with other methods with similar backbones under the {\em 1-clip and center crop setting}.  ``-'' indicates the numbers are not available for us. \textcolor{red}{\footnotesize{$^1$}} Pre-trained on Sports1M. }
\label{tbl:sota_sth}
\vspace{-3mm}
\end{table}

\vspace{-4mm}
\paragraph{\bf Study on long-term TDM.} We employ short-term TDM for the early stages, and compare with different forms of long-term TDM (L-TDM) placed on the latter stages. The results are reported in Table~\ref{tbl:ltdm}. For L-TDM design, we first compare with two baseline architecture: (1) no attention modeling in Eq.~(\ref{equ:ltdm}) and directly adding the difference representation into frame-level features; (2) channel attention modeling just as TEA~\cite{li2020tea}. It is observed that our proposed spatiotemporal attention form of L-TDM is better than no attention (52.3\% vs.44.1\%) and channel attention (52.3\% vs. 50.9\%). Then, we investigate the effectiveness of multi-scale architecture in difference feature extraction and it is able to improve performance from 49.7\% to 52.3\%, which confirms its effectiveness of large receptive field for difference feature extraction. Finally, we compare the performance of bidirectional difference with one-directional difference, and it helps to improve performance by 2.3\%.

\vspace{-4mm}
\paragraph{\bf Study on the location of S-TDM and L-TDM .} We perform the ablation study on which stage to use short-term TDM (S-TDM) or long-term TDM (L-TDM). The results are shown in Table~\ref{tbl:ablation_loc}. From these results, we see that adding more S-TDMs into the main network will increase the network computational cost slightly due to its feature extraction for temporal difference representation. The setting of using S-TDM in stages 1-2 and L-TDM in stages 3-5 obtains the best recognition accuracy and the computational cost is also reasonable.

\begin{table}[t]
\centering
\small
\tablestyle{1.6pt}{1}
\begin{tabular}{l|c|c|c|c|c}
\multicolumn{1}{c|}{\bfseries Method} & \bfseries Backbone  &\multicolumn{1}{c|}{\bfseries Frames} &  \bfseries GFLOPs  & \bfseries Top1  & \bfseries Top5  \\
\shline
TSN~\cite{DBLP:conf/eccv/WangXW0LTG16} &    InceptionV3 &    25$\times$1$\times$10  &    3.2$\times$250 &    72.5 &   90.2 \\
S3D-G~\cite{DBLP:conf/eccv/XieSHTM18} & InceptionV1 & 64$\times$10$\times$3 & 71.4$\times$30 & 74.7 & 93.4 \\
R(2+1)D~\cite{DBLP:conf/cvpr/TranWTRLP18} & ResNet34 & 32$\times$10$\times$1 & 152$\times$10 & 74.3 & 91.4\\
TSM~\cite{DBLP:conf/iccv/LinGH19} & ResNet50 & 16$\times$10$\times$3 & 65$\times$30 & 74.7 & 91.4\\
TEINet~\cite{DBLP:journals/corr/abs-1911-09435} & ResNet50 & 16$\times$10$\times$3 & 66$\times$30 & 76.2 & 92.5 \\
TEA~\cite{li2020tea} & ResNet50 & 16$\times$10$\times$3 & 70$\times$30& 76.1 & 92.5 \\
TAM~\cite{tam} & bLResNet50 & 48$\times$3$\times$3 & 93.4$\times$9 & 73.5 & 91.2 \\
TANet~\cite{TANet} & ResNet50 & 16$\times$4$\times$3 &  86$\times$12 & 76.9 & 92.9 \\
\hline \hline 
ARTNet~\cite{DBLP:conf/cvpr/WangL0G18} & ResNet18 & 16 $\times$25$\times$10 & 23.5$\times$250 & 70.7 & 89.3 \\
I3D~\cite{I3D} & InceptionV1 & 64$\times$N/A$\times$N/A & 108$\times$N/A & 72.1 & 90.3\\
NL I3D~\cite{nonlocal} & ResNet50 & 128$\times$10$\times$3 & 282$\times$30 & 76.5 & 92.6 \\
SlowOnly~\cite{DBLP:journals/corr/abs-1812-03982} & ResNet50 & 8$\times$10$\times$3 & 41.9$\times$30 & 74.8 & 91.6 \\
SlowFast~\cite{DBLP:journals/corr/abs-1812-03982} & ResNet50 & (4+32)$\times$10$\times$3 & 36.1$\times$30 & 75.6 & 92.1 \\
SlowFast~\cite{DBLP:journals/corr/abs-1812-03982} & ResNet50 & (8+32)$\times$10$\times$3 & 65.7$\times$30 & 77.0 & 92.6 \\
SmallBigNet~\cite{smallbignet}& ResNet50 & 8$\times$10$\times$3 & 57$\times$30 & 76.3 & 92.5 \\
CorrNet~\cite{CorrNet}& ResNet50 & 32$\times$10$\times$1 & 115$\times$10 & 77.2 & - \\
\hline \hline 
TDN (Ours) & ResNet50& 8$\times$10$\times$3 & 36$\times$30 & 76.6 & 92.8 \\
TDN (Ours) & ResNet50& 16$\times$10$\times$3 & 72$\times$30 & 77.5 & 93.2 \\
TDN (Ours) & ResNet50& (8+16)$\times$10$\times$3 & 108$\times$30 & {\bf 78.4} & {\bf 93.6} \\
\hline \hline 
NL I3D~\cite{nonlocal} & ResNet101 & 128$\times$10$\times$3 & 359$\times$30 & 77.7 & 93.3 \\
ip-CSN~\cite{csn} & ResNet101 & 32$\times$10$\times$3 & 83.0$\times$30 & 76.7 & 92.3 \\
SlowFast~\cite{DBLP:journals/corr/abs-1812-03982} & ResNet101 & (8+32)$\times$10$\times$3 & 106$\times$30 & 77.9 & 93.2 \\
SlowFast~\cite{DBLP:journals/corr/abs-1812-03982} & ResNet101 & (16+64)$\times$10$\times$3 & 213$\times$30 & 78.9 & 93.5 \\
SmallBigNet~\cite{smallbignet}& ResNet101 & 32$\times$4$\times$3 & 418$\times$12 & 77.4 & 93.3 \\
CorrNet~\cite{CorrNet}& ResNet101 & 32$\times$10$\times$3 & 224$\times$30 & 79.2 & - \\
\hline \hline 
TDN (Ours) & ResNet101& 8$\times$10$\times$3 & 66$\times$30 & 77.5 & 93.6 \\
TDN (Ours) & ResNet101& 16$\times$10$\times$3 & 132$\times$30 & 78.5 & 93.9 \\
TDN (Ours) & ResNet101& (8+16)$\times$10$\times$3 & 198$\times$30 & {\bf 79.4} & {\bf 94.4} \\
\hline \hline 
SlowFast~\cite{DBLP:journals/corr/abs-1812-03982} & R101+NL & (16+64)$\times$10$\times$3 & 234$\times$30 & {\bf 79.8} & 93.9 \\
X3D~\cite{x3d} & X3D-XL & 16$\times$10$\times$3 & 48.4$\times$30 & 79.1 & 93.9 \\
\end{tabular}
\vspace{0.5mm}
\caption{\small {\bf Comparison with the state-of-the-art methods on the validation set of  Kinetics-400}. We instantiate our TDN with the backbones of ResNet50 and ResNet101. For fair comparison, we compare with the other methods by using the similar backbones without pre-training on extra videos. ``-'' indicates the numbers are not available for us.}
\label{tbl:sota_k400}
\vspace{-3mm}
\end{table}

\vspace{-3mm}
\paragraph{\bf Short-term vs. long-term modeling.} We conduct comparative study to separately investigate the effectiveness of S-TDM and L-TDM. The results are summarized in Table~\ref{tbl:comp1}. We first report the performance of baseline without S-TDM or L-TDM, namely only with 1D temporal convolutions in latter stages for temporal modeling, and its accuracy is 46.6\%. Then we separately add S-TDM and L-TDM into the baseline, and they obtain the performance of 51.5\% and 49.9\%. The superior performance of S-TDM to L-TDM might be ascribed to the fact that local motion information is crucial for action recognition. Finally, combining S-TDM and L-TDM could boost performance to 52.3\%, which implies the complementarity of two modules.

\vspace{-3mm}
\paragraph{\bf Comparison with other temporal modules.} Finally, we compare our proposed TDM with other temporal modeling methods, and the results are reported in Table~\ref{tbl:modulecmp}. We compare our TDM with three temporal modules: temporal convolution~\cite{DBLP:conf/cvpr/TranWTRLP18}, TSM~\cite{DBLP:conf/iccv/LinGH19}, and TEINet~\cite{DBLP:journals/corr/abs-1911-09435}. First, these methods all use the ResNet50 as backbones and 8 frames as input. In this setting, their FLOPs are similar to our TDN. We find that the performance of our TDN is much better than those baselines with similar FLOPs, demonstrating the effectiveness of explicit temporal difference operation. Then, for fair comparison, we also implement other temporal modules taking the same number of frames as ours (i.e., 40 frames denoted by ++), and we observe that simply inputting more frames will not contribute much to improve recognition accuracy. We analyze that these temporal modules still lack sufficient modeling capacity to well capture fine-grained motion information and thus more frames will make them over-fit with appearance more seriously. On the other hand, thanks to temporal difference operation, our TDM is able to focus more on the motion information and thus improve the recognition accuracy with more frames.

\subsection{Comparison with the state of the art}
\label{sec:cmp}

After the ablation study of 8-frame TDN on Something-Somthing V1 dataset, we directly transfer its optimal setting to the datasets of Something-Something V2 and Kinetics-400. In this subsection, we compare our TDN with those state-of-the-art methods on these benchmarks. As expected, sampling more frames can further improve the accuracy, but also increases the FLOPs. We report the performance of both 8-frame TDN and 16-frame TDN. For fair comparison, we simply list the performance of methods solely using RGB without pre-training on extra video datasets. 

The results are summarized in Table ~\ref{tbl:sota_sth} and Table~\ref{tbl:sota_k400}. For fair comparison with previous methods, we use 1 clip and center crop testing scheme on the Something-Something dataset and 10 clips and 3 crops for testing on the Kinetics-400 dataset. We first compare with 2D CNN based baselines with late fusion for long-range temporal modeling such as TSN~\cite{DBLP:conf/eccv/WangXW0LTG16} and TRN~\cite{DBLP:conf/eccv/ZhouAOT18}, and see that our TDN outperforms these baseline methods significantly on both datasets. Then, we compare our TDN with 2D CNN with temporal modules for all stages, such as S3D~\cite{DBLP:conf/eccv/XieSHTM18}, R(2+1)D~\cite{DBLP:conf/cvpr/TranWTRLP18}, TSM~\cite{DBLP:conf/iccv/LinGH19}, TEINet~\cite{DBLP:journals/corr/abs-1911-09435}, TANet~\cite{TANet}, TAM~\cite{tam}, and GSM~\cite{GSM}, and our TDN consistently outperforms them on both datasets, demonstrating the effectiveness of TDM in temporal modeling for action recognition. After this, we compare with more recent 3D CNNs based methods, such as I3D~\cite{I3D}, Non-local I3D~\cite{nonlocal}, and SlowFast~\cite{DBLP:journals/corr/abs-1812-03982}, and our TDN can still obtain slightly better performance than those methods, with a relatively smaller computational cost. Finally, we compare more recent video recognition networks, such as SmallBigNet~\cite{smallbignet}, V4D~\cite{zhang2019v4d}, CorrelationNet~\cite{CorrNet}, and X3D~\cite{x3d}. Our best result significantly outperforms previous methods on Something-Something V1 and is on par with the previous best performance on the Kinetics dataset. The best performance on the Kinetics dataset is the combination of SlowFast and Non-local Net, which is slightly better than ours for Top1 accuracy yet with lower Top5 accuracy and higher FLOPs.

\begin{figure}[t]
    \centering
    \includegraphics[width=0.48\textwidth]{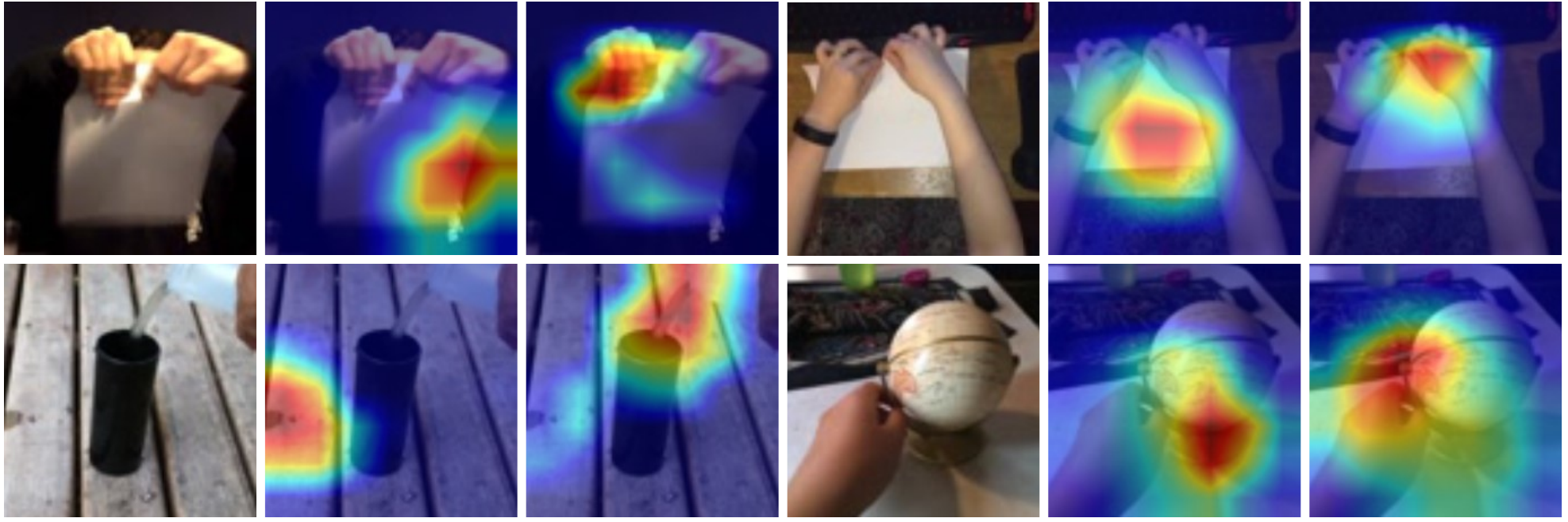}
    \caption{Visualization of activation maps with CAM. {\em Left}: video, {\em Middle}: baseline, {\em Right}: TDN. In this visualization, we train a 8-frame network with TDMs (TDN) or temporal convolutions (Baseline). For simplicity, we only visualize the CAM on the center frames. More visualization examples on 8 frames could be found in supplementary material.}
    \label{fig:visualizaiton}
    \vspace{-2mm}
\end{figure}

\subsection{Visualization of activation maps}

We visualize the class activation maps with Grad-CAM~\cite{ZhouKLOT16,selvaraju2020grad} and results are shown in Figure~\ref{fig:visualizaiton}. In this visualization, we take 8 frames as input and only plot the activation maps in the center frames. These visualization results indicate that baseline with only temporal convolutions fails to focus on motion-salient regions, while our TDN is able to localize more action-relevant regions, thanks to our proposed TDMs for short-term and long-term temporal modeling. For example, our TDN pays more attention to the hand motion with interaction objects, while the temporal convolution may only focus on the background. More visualization examples and analysis could be found in the supplementary material.

\section{Conclusion}

In this paper, we have presented a new video-level framework, termed as TDN, for learning action models from the entire video. The core contribution of TDN is to generalize temporal difference operator into efficient and general temporal modules (TDM) with specific designs, for capturing both short-term and long-term temporal information in a video. We present two customized forms for the implementation of TDMs and systematically assess their effects on temporal modeling. As demonstrated on the Kinetics-400 and Something-Something dataset, our TDN is able to yield superior performance to previous state-of-the-art methods of using similar backbones. 

In addition, we present an in-depth ablation study on TDMs to investigate the effect of temporal difference operation, and demonstrate that it is more effective to extract fine-grained temporal information than a standard 3D convolution with more frames. We hope our analysis provides more insights about temporal difference operation, and TDM might provide an alternative to 3D convolution for temporal modeling in videos.

{\small \paragraph{\bf Acknowledgements.} This work is supported by the National Science Foundation of China (No. 62076119, No. 61921006), Program for Innovative Talents and Entrepreneur in Jiangsu Province, and Collaborative Innovation Center of Novel Software Technology and Industrialization.}


\section*{Appendix}
This appendix contains the results on the UCF101 and HMDB51, running time analysis of TDN, and more visualization results.

\appendix
\section{Results on the UCF101 and HMDB51}
\begin{table}[h]
\centering
\small
\tablestyle{1.6pt}{1}
\begin{tabular}{l|c|c|c|c}
\bf Method & \bf Pretrain &\bf Backbone & \bf UCF101 & \bf HMDB51 \\
\shline
 TSN~\cite{DBLP:conf/eccv/WangXW0LTG16} & ImageNet & Inception V2 & 86.4\% & 53.7\% \\
 P3D~\cite{Qiu2017LearningSR} & ImageNet & ResNet50 & 88.6\% & - \\
 C3D~\cite{DBLP:conf/iccv/TranBFTP15} & Sports-1M & ResNet18 & 85.8\% & 54.9\% \\
 I3D~\cite{I3D} & ImageNet+Kinetics & Inception V2 & 95.6\% & 74.8\% \\
 ARTNet~\cite{DBLP:conf/cvpr/WangL0G18} & Kinetics & ResNet18 & 94.3\% & 70.9\% \\
 S3D~\cite{DBLP:conf/eccv/XieSHTM18} & ImageNet+Kinetics & Inception V2 & 96.8\% & 75.9\% \\
 R(2+1)D~\cite{DBLP:conf/cvpr/TranWTRLP18} & Kinetics & ResNet34 & 96.8\% & 74.5\% \\
 TSM~\cite{DBLP:conf/iccv/LinGH19} & Kinetics & ResNet50 & 96.0\% & 73.2\% \\
 STM~\cite{DBLP:conf/iccv/JiangWGWY19} &  ImageNet + Kinetics & ResNet50 & 96.2\% & 72.2\% \\
 TEA~\cite{li2020tea} & ImageNet + Kinetics & ResNet50 & 96.9\% & 73.3\% \\
\hline 
TDN(Ours) & ImageNet + Kinetics & ResNet50 & {\bf 97.4\%} & {\bf 76.3\%} \\
\end{tabular}
\vspace{1mm}
\caption{Comparison with the state-of-the-art methods on {\bf UCF101 and HMDB51}. }
\label{tbl:sota}
\end{table}
To further verify the generalization ability of TDN, we transfer the learned 16-frame TDN models from the Kinetics-400 dataset to the UCF101 and HMDB51. These two datasets are relatively small and the action recognition performance on them already saturates. We follow the standard evaluation scheme on these two datasets and report the mean accuracy over three splits. The results are summarized in Table~\ref{tbl:sota}. We compare our TDN with previous state-of-the-art methods such as 2D baselines of TSN~\cite{DBLP:conf/eccv/WangXW0LTG16}, 3D CNNs of I3D~\cite{I3D} and C3D~\cite{DBLP:conf/iccv/TranBFTP15}, R(2+1)D~\cite{DBLP:conf/cvpr/TranWTRLP18}, and other temporal modeling methods~\cite{li2020tea,DBLP:conf/iccv/JiangWGWY19}. From the results, we can see that our TDN is able to outperform these methods, and the performance improvement is more evident on the dataset of HMDB51 by around 2.5\%. The action classes in HMDB51 are more relevant with motion information, and thus temporal modeling is more important on this dataset.

\section{Running time analysis}

We report the inference time of our TDN with on Tesla V100 as follows. The testing batchsize is set as 16 and the running time include all evaluation, including loading data and network inference. The results are reported in Table~\ref{tbl:time}. From these results, we  see that our TDN is slower than previous method but still could run in real-time (i.e. $\geq$25 FPS).

\begin{table}[t]
\centering
\small
\tablestyle{1.6pt}{1.05}
\begin{tabular}{c|c|c|c}

\bf Method & \bf Frames$\times$Clips$\times$Crops & \bf Time (ms/video) & \bf Top1 (\%) \\
\shline
TSN~\cite{DBLP:conf/eccv/WangXW0LTG16} &  $8 \times 1 \times 1$ & 7.9 & 19.7 \\
TSM~\cite{DBLP:conf/iccv/LinGH19} &  $16 \times 1 \times 1$ & 16.7 & 47.2 \\
STM~\cite{DBLP:conf/iccv/JiangWGWY19} &  $8 \times 1 \times 1$ & 11.1 & 47.5 \\
I3D~\cite{I3D} &  $32 \times 3 \times 2$ & 2095 & 41.6 \\
\hline
S-TDM & $8 \times 1 \times 1$ & 12.3 & 49.5 \\
L-TDM & $8 \times 1 \times 1$ & 15.8 & 48.9 \\
TDN & $8 \times 1 \times 1$ & 22.1 & 52.3 \\
\end{tabular}
\vspace{1mm}
\caption{Running time analysis on a Tesla V100.}
\label{tbl:time}
\end{table}

\section{Visualization analysis}

\begin{figure*}
    \centering
    \includegraphics[width=0.8\textwidth]{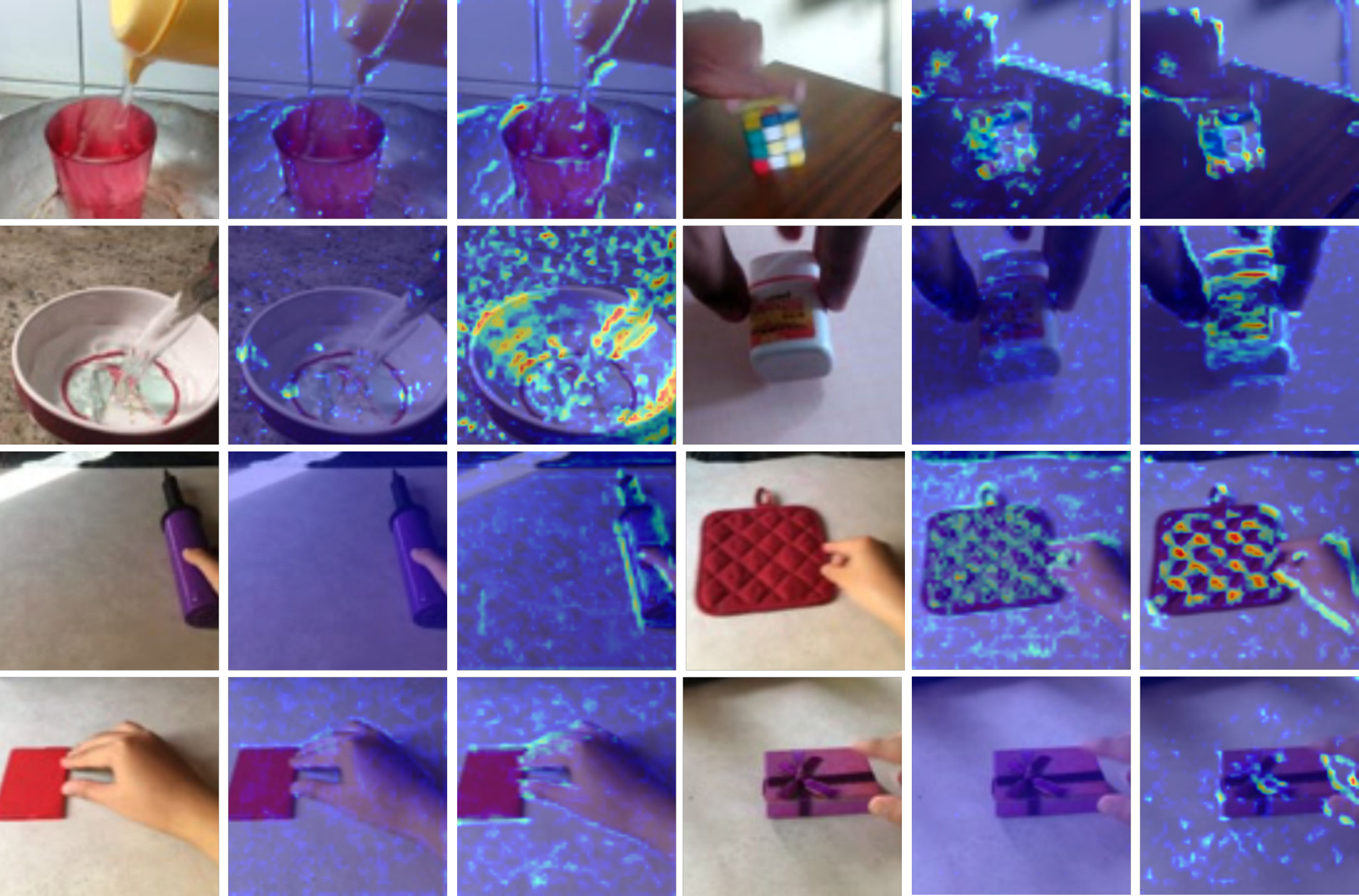}
    \caption{Visualization of Res2 features with Grad-CAM. We use 8-frame TDN models to visualize on the Something-Something V1 dataset. Left: video, Middle: baseline, Right: TDN with S-TDM. Note that we only show visualization on the center frame of sampled 8 frames.}
    \label{fig:visualizaiton1}
\end{figure*}

\begin{figure*}
    \centering
    \includegraphics[width=0.8\textwidth]{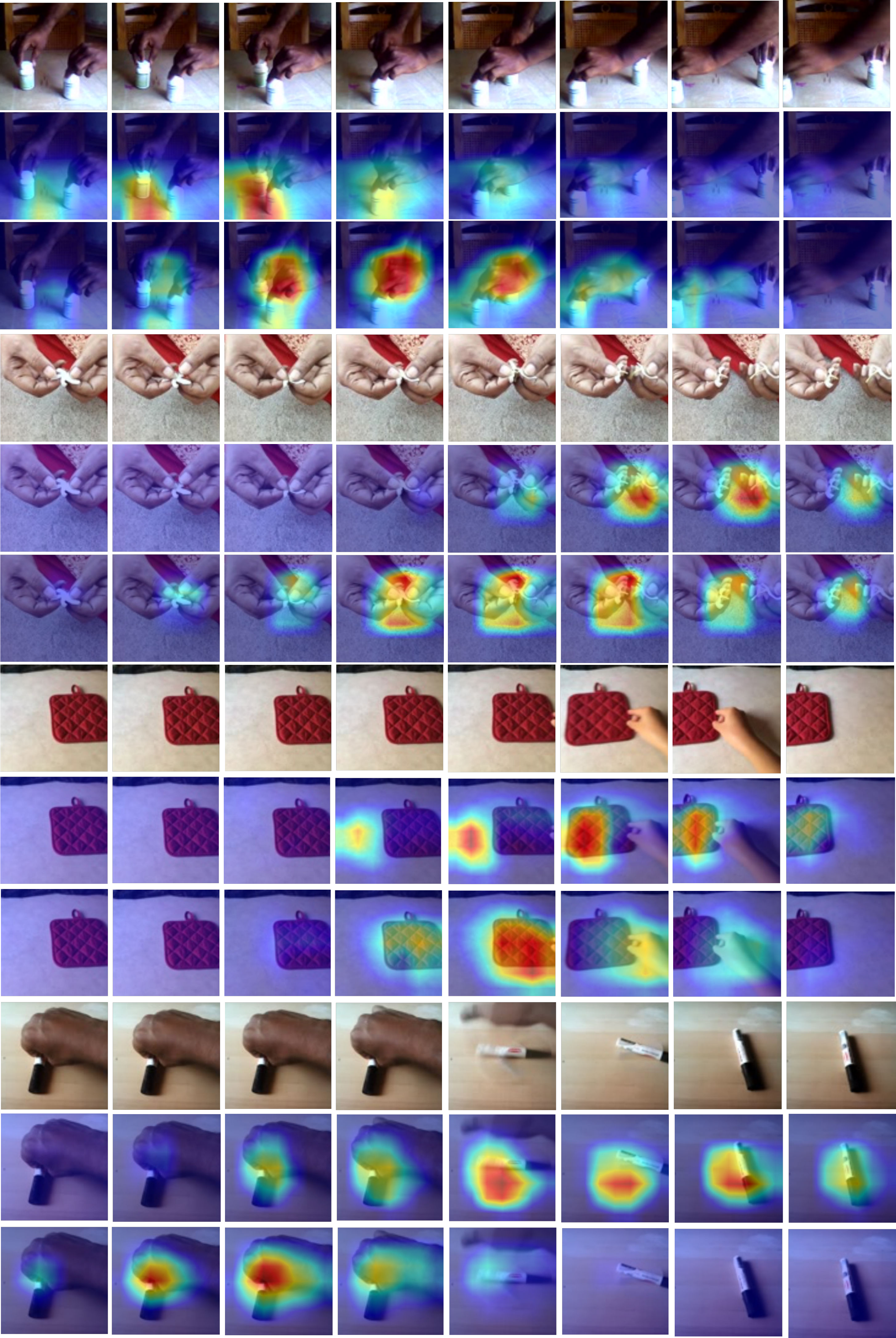}
    \caption{Visualization of activation maps with Grad-CAM. We use 8-frame TDN models to visualize on the Something-Something V1 dataset. In the first row, we plot the 8 RGB frames. In the second row, we plot the activation maps of the baseline method without temporal difference module (TDM). In the third row, we plot the activation maps of the TDN models.}
    \label{fig:visualizaiton2}
\end{figure*}

\begin{figure*}
    \centering
    \includegraphics[width=0.8\textwidth]{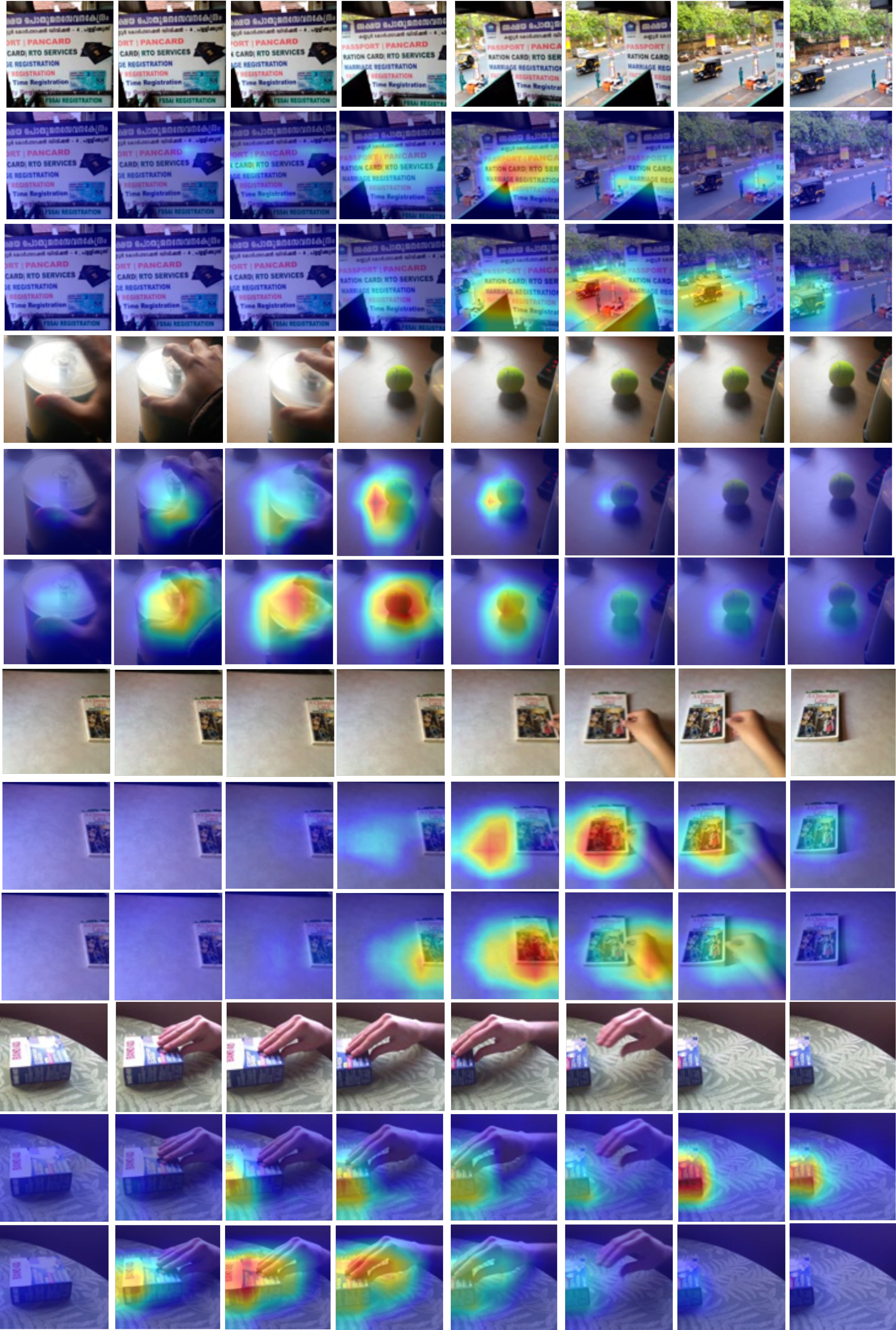}
    \caption{Visualization of activation maps with Grad-CAM. We use 8-frame TDN models to visualize on the Something-Something V1 dataset. In the first row, we plot the 8 RGB frames. In the second row, we plot the activation maps of the baseline method without temporal difference module (TDM). In the third row, we plot the activation maps of the TDN models.}
    \label{fig:visualizaiton3}
\end{figure*}

To further investigate the performance the TDN models, we use the technique of Grad-CAM~\cite{selvaraju2020grad} to visualize the feature representation of different models. Specifically, to better understand the effect of short-term TDM, we visualize the the features in Res2 stage of baseline model (corresponding to the first row in Table 1(e) of main article) and the TDM model only with S-TDM (corresponding to third row in in Table 1(e) of main article), and the results are shown in Figure~\ref{fig:visualizaiton1}. Note that, these visualizations only are performed on the center frame of 8-frame models. From these results, the models equipped with S-TDM focuses more on motion-relevant information. Then, we give more visualization examples of activation maps in Figure~\ref{fig:visualizaiton2} and Figure~\ref{fig:visualizaiton3}. In these results, we give the visualization results on 8 frames and compare our TDM models with the baseline method (corresponding to the first row in Table 1(e) of main article). We could see that our TDN is able to yield more reasonable class activation maps than the baseline method.

{\small
\bibliographystyle{ieee_fullname}
\bibliography{TDN}
}

\end{document}